\documentclass[journal]{IEEEtran}

\ifCLASSINFOpdf
\else
   \usepackage[dvips]{graphicx}
\fi
\usepackage{url}

\usepackage[mathletters]{ucs}
\usepackage[utf8x]{inputenc}

\usepackage{float}
\usepackage{graphicx, cite}
\usepackage{multirow, booktabs, makecell}
\usepackage[table]{xcolor}
\definecolor{gray}{rgb}{0.85,0.85,0.85}
\usepackage{footnote}
\makesavenoteenv{tabular}
\makesavenoteenv{table}

\usepackage{amsmath}
\usepackage{amsfonts}
\usepackage{amssymb}
\usepackage{amsthm}
\usepackage{braket}
\usepackage[scientific-notation=true]{siunitx}

\newcommand\percentage[2][round-precision = 2]{
    \SI[round-mode = places,
        scientific-notation = fixed, fixed-exponent = 0,
        output-decimal-marker={.}, #1]{#2e2}{\percent}
}

\newcommand\boldpercentage[2][round-precision = 2]{
    \SI[round-mode = places,
        scientific-notation = fixed, fixed-exponent = 0,
        output-decimal-marker={.}, #1]{#2e2}{\boldsymbol{\percent}}
}

\newcommand{\card}[1]{\left| #1 \right|}
\newcommand{\simPG}[1][x]{\sim \frac{\mathcal{P}(a #1)}{a} + \mathcal{N}(0,b^2)}

\begin{document}

\title{PoGaIN: Poisson-Gaussian Image Noise\\ Modeling from Paired Samples}

\author{\IEEEauthorblockN{Nicolas Bähler$^*$, Majed El Helou$^*$, Étienne Objois, Kaan Okumuş, and Sabine Süsstrunk, \IEEEmembership{Fellow, IEEE}.}

    \thanks{$^*$ Both authors have equal contributions.}
    \thanks{Submitted for review on October 10th, 2022, revised on November 20th, 2022, and accepted on November 26th, 2022}
    \thanks{Work carried out in the Image and Visual Representation Lab (IVRL) at the School of Computer and Communication Sciences, EPFL, Switzerland. \{nicolas.bahler, sabine.susstrunk\}@epfl.ch, melhelou@ethz.ch.\\
        Corresponding author: Nicolas Bähler}
}

\maketitle
\IEEEpeerreviewmaketitle

\begin{abstract}
    Image noise can often be accurately fitted to a Poisson-Gaussian distribution. However, estimating the distribution parameters from a noisy image only is a challenging task. Here, we study the case when paired noisy and noise-free samples are accessible. No method is currently available to exploit the noise-free information, which may help to achieve more accurate estimations. To fill this gap, we derive a novel, cumulant-based, approach for Poisson-Gaussian noise modeling from paired image samples. We show its improved performance over different baselines, with special emphasis on MSE, effect of outliers, image dependence, and bias. We additionally derive the log-likelihood function for further insights and discuss real-world applicability.
\end{abstract}


{\let\thefootnote\relax\footnotetext{All code and supplementary material at \url{https://github.com/IVRL/PoGaIN}}}

\section{Introduction} \label{sec:intro}
Noise always affects image capture in any imaging pipeline. Modeling noise distribution is thus crucial for analyzing imaging devices, datasets~\cite{zhang2019poisson,zhou2020w2s}, and developing denoising methods, especially blind ones~\cite{elhelou2020blind,tran2020gan,6607209,7669699}. Those approaches include noise parameter estimation prior to the noise reduction, and hence do not rely on the noise level being known. Other learning-based techniques even go a step further and are noise model-blind, meaning that no fixed noise model is imposed~\cite{8954460, 10.1007/978-3-030-31723-2_21}. Here, we assume a common noise model, the Poisson-Gaussian noise model, composed of a shot and a read noise component. The former is modeled with a Poisson distribution, emerging from the particle nature of light whose intensity the sensor estimates over a finite duration of time. The latter is modeled with a Gaussian distribution, notably for raw images that are processed by the different steps in the image processing pipeline, which can modify the distribution.

In their seminal work, Foi~\emph{et al.}~\cite{foi2008practical} also propose a Poisson-Gaussian model for the noise distribution. Further, the authors introduce a clever solution for fitting the noise model parameters from a noisy input image. Their algorithm begins with local expectation and standard deviation estimates from image parts that are assumed to depict a single underlying intensity value. The global parametric model is then fitted through a maximum likelihood search based on the local estimates. Multiple assumptions are made in order to reach a final estimate, in part due to the lack of input information beyond the noisy image. Our premise is that when modeling datasets or analyzing an imaging system, we may be able to acquire paired noisy and noise-free images. We exploit this additional information and study the problem of modeling noise \textit{with paired samples}.

We propose a novel method that estimates the parameters of the aforementioned noise model based on noisy and noise-free image pairs that can be used to develop new blind denoising algorithms. The additional information of the noise-free version of a given image enables our approach to significantly outperform the method introduced by Foi~\emph{et al.}~\cite{foi2008practical}. We also train a neural network based noise model estimator and show that we in addition outperform this learning-based alternative. Finally, for the sake of comparison, we introduce a variance-based baseline method, which also takes advantage of noisy and noise-free image pairs.

\section{Related work}\label{sec:related}
Denoising is one of the most fundamental tasks in image restoration, with both
theoretical impact and practical applications. Most classic denoisers, for
instance PURE-LET~\cite{luisier2011image}, KSVD~\cite{KSVD}, WNNM~\cite{WNNM},
BM3D~\cite{BM3D}, and EPLL~\cite{EPLL}, require knowledge of the noise level in
the input test image. Deep learning image denoisers that have shown improved
empirical performance~\cite{huang2022neighbor2neighbor,ma2021deep} also require
knowledge of noise distributions, if not at test time~\cite{ffdnet}, then at
least for training~\cite{elhelou2020blind,elhelou2022bigprior}. This is due to
the degradation overfitting of deep neural networks~\cite{el2020stochastic}.
Noise modeling is thus important for denoisers at test time, but also for
acquisition system analysis and dataset modeling for training these denoisers.
Past research has focused on modeling noise from noisy images without relying on
ground truth, i.e., noise-free, information~\cite{foi2008practical}. Interesting
approaches, for example Sparse Modeling~\cite{7990585}, Dictionary
Learning~\cite{2018} or non-local image denoising methods like
SAFPI~\cite{10.1371/journal.pone.0208503}, have been developed to push overall
denoising performance. However, none of these methods allow easy use of
noise-free data when it is available. For Poisson-Gaussian noise modeling, for
example, both FMD~\cite{zhang2019poisson} and W2S~\cite{zhou2020w2s} rely on a
noise modeling method that does not consider ground truth noise-free
images~\cite{foi2008practical}. Hence, our approach to model the
\textbf{Po}isson-\textbf{Ga}ussian \textbf{I}mage \textbf{N}oise (PoGaIN)
distribution exploits paired samples (noisy and noise-free images), which
significantly improves the modeling accuracy. Our method is based on the
cumulant expansion, which is also used by other authors to derive estimators for
PoGaIN model parameters, but for different input types, such as noisy image time
series~\cite{6235897} or single noisy images~\cite{zhang:pastel-00003273}.

\section{Mathematical formulation}\label{sec:math}
\subsection{Poisson-Gaussian model}\label{subsec:model}
The Poisson-Gaussian noise model proposed by Foi~\emph{et al.}~\cite{foi2008practical} consists of two components, the Poisson shot noise and the Gaussian read noise, which are assumed to be independent.
The signal-dependent Poisson component $\eta_p$ and signal-independent Gaussian component $\eta_g$ are defined, respectively, by
\begin{equation}
    \eta_p = \frac{1}{a} \alpha, \quad \eta_g = \beta, \quad \alpha \sim \mathcal{P}(ax), \quad \beta \sim \mathcal{N}(0,b^2),
\end{equation}
where $x$ is the ground truth (noise-free) signal, and $a$ and $b$ are distribution parameters. The complete model is made up of the sum of these two components
\begin{equation}
    y = \eta_p + \eta_g = \frac{1}{a} \alpha + \beta,
    \label{eq:poisgaus}
\end{equation}
where $y$ is the observed (noisy) signal. We note to the reader that the $a$ and $b$ in~\cite{foi2008practical} correspond to our $a^{-1}$ and our $b^2$, respectively. Thus, our $a$ is equal to the quantum efficiency in percent.

As derived in the supplementary material, the following properties hold for $\eta_p$
\begin{equation}
    \mathbb{E}[\eta_p] = x, \quad \mathbb{V}[\eta_p] = \frac{x}{a},
\end{equation}
which shows that the Poisson component is indeed signal-dependent and that the Gaussian component, having constant mean and variance, is signal independent.
Consequently, we derive the expected value and variance of the observation $y$
\begin{equation}
    \mathbb{E}[y]=x, \quad \mathbb{V}[y] = \frac{x}{a} + b^2.
    \label{eq:statpoisgaus}
\end{equation}

\subsection{Likelihood derivation}\label{subsec:likelihood}
The noise model, presented in Equation~\eqref{eq:poisgaus}, leads to the following expression for the likelihood

\begin{equation}
    \mathcal{L}(y|a,b,x) =\prod_{i}\sum_{k=0}^\infty\frac{(ax_i)^k}{k!b\sqrt{2\pi}}\exp{\left(-ax_i-\frac{(y_i-k/a)^2}{2b^2}\right)},
    \label{eq:likelihood_func}
\end{equation}

where $y$ is the captured noisy image, $x$ is the ground truth noise-free image and $i$ is the pixel index in the vectorized representation of an image. The complete derivation of the likelihood function is given in the supplementary material.

The noise parameters $\hat{a}$ and $\hat{b}$ that optimize for the maximum likelihood are then given by
\begin{equation}
    \begin{split}
        \hat{a}, \hat{b} = \arg \max_{a, b} \mathcal{L}(y|a,b,x).
    \end{split}
    \label{mlsoln}
\end{equation}

Optimizing over the log-likelihood $\mathcal{LL}$ is computationally inefficient. To improve convergence, we truncate the summation over $k$ to a $k_{max}$ such that most of the weight of the sum lies in $k$ values below $k_{max}$ (details in code). Nonetheless, optimizing $\mathcal{LL}$ is not a viable solution to our problem. However, as we show in our Analysis~\ref{sec:loglikelihood}, the log-likelihood can still provide empirical insight.

\section{Proposed method}\label{sec:method}
Rather than interpreting $x$ and $y$ as two observed images, we consider $x$ and $y$ as a collection of samples from two distributions $\mathcal{X}$ and $\mathcal{Y}$ and define a random variable $X \sim \mathcal{X}$ such that
\begin{equation}
    \mathbb{P} [X = x_i] = \frac{\card{\set{k : x_k = x_i}}}{n},
\end{equation}
where $n$ corresponds to the number of samples (i.e., the number of pixels in $x$ and $y$). We define $\mathcal{Y}$ to be the distribution of the Poisson-Gaussian noise model over the distribution $\mathcal{X}$. Formally, introducing another random variable $Y \sim \mathcal{Y}$, we get
\begin{equation}
    Y \sim \mathcal{Y} = \frac{\mathcal{P}(a \mathcal{X})}{a} + \mathcal{N}(0,b^2).
\end{equation}
Next, we obtain the $2$-nd and $3$-rd cumulant of $\mathcal{Y}$ as a system of equations
\begin{equation}
    \label{eq:cumu}
    \begin{split}
        \left\{
        \begin{array}{l}
            \kappa_2[\mathcal{Y}] = \frac{\overline{x}}{a} + \overline{x^2} - \overline{x}^2 + b^2 \\
            \kappa_3[\mathcal{Y}] = \overline{x^3} - 3 \overline{x^2}\overline{x} + 2 \overline{x}^3 + 3\frac{\overline{x^2}}{a} - 3 \frac{\overline{x}^2}{a} + \frac{\overline{x}}{a^2}
        \end{array}
        \right.,
    \end{split}
\end{equation}
where $\overline{x}$ denotes the mean of $x$ (for example $\overline{x^k}^j = (\frac{1}{n}\sum_i x_i^k)^j$). Both $\kappa_2[\mathcal{Y}]$ and $\kappa_3[\mathcal{Y}]$ can be estimated with an unbiased estimator  (k-statistic). Therefore, Equations~\eqref{eq:cumu} form a system of two equations, with two unknowns, $a$ and $b$, which is solved by our cumulant-based method (\textbf{\textit{OURS}}).\\

\section{Experimental evaluation} \label{sec:experiments}

\subsection{Data processing}
The dataset we use is based on the Berkeley Segmentation
Dataset 300~\cite{dataset:MartinFTM01}. We synthesize noise stochastically by picking a seed $s \in \set {0, \dots, 9}$, $a \in [1, 100]$ and $b \in [0.01,0.15]$ and distort images from the training set~\cite{dataset:MartinFTM01} with it, resulting in noisy and noise-free image pairs. For validation, we pick $10$ images out of the test set of~\cite{dataset:MartinFTM01}. For each seed $s \in \set {0, \dots, 9}$, and for $25$ linearly spaced values for $a \in [1, 100]$ and $b \in [0.01,0.15]$, we synthesize an image pair, resulting in a total of $62500$ pairs for evaluation.

\subsection{Baseline methods}
\subsubsection{\textit{FOI}}
The method proposed by Foi~\emph{et al.}~\cite{foi2008practical} estimates $a$ and $b$ by segmenting pixels assumed to have the same underlying value but to be distorted by noise into non-overlapping intensity level sets. Further, a local estimation of multiple expectation and standard-deviation pairs is carried out. Finally, a global parametric model fitting using those local estimates is performed. \textit{FOI} only uses $y$, and does not provide a way to exploit $x$ even if it is available. Naturally, that makes a good estimation of $a$ and $b$ more challenging.

\subsubsection{\textit{CNN}}
For the sake of comparison, we design a convolutional neural network (CNN) that we train to predict $a$ and $b$. The detailed architecture of the CNN is described in our code repository. The CNN takes only the noisy image $y$ as input. We cannot rule out that a more complex neural network based solution could outperform this CNN. We provide it as an additional baseline, and it is not the focus of this article.

\subsubsection{\textit{VAR}}
For fairness of comparison, we design a baseline method that also takes advantage of noisy and noise-free image pairs. To derive it, we define $Y_i = \set{y_j : x_j = x_i}$ as the set of all pixels of $y$ for which the corresponding pixels in $x$ have the same intensity as $x_i$. This approach is based on the variance across the pixel sets $Y_i$.
First, the theoretical mean of $Y_i$ is $x_i$, and hence we can compute the empirical variance of $Y_i$
\begin{equation}
    \label{eq:var_emp_var}
    \mathbb{V} [Y_i] = \frac{1}{\card{Y_i}} \sum_{y_k \in Y_i} (y_k - x_i)^2.
\end{equation}
Additionally, $y_j \simPG[x_i]$. Thus, according to Equation~\eqref{eq:statpoisgaus}, $\mathbb{V} [Y_i] \approx \frac{x_i}{a} + b^2$. Using this observation, we select $a, b$ such that
the above approximation holds as closely as possible
between the two values for any given $i$, where $\mathbb{V} [Y_i]$ is computed using Equation~\eqref{eq:var_emp_var}. Finally, we obtain an estimation of $a,b$ by computing
\begin{equation}
    \hat a, \hat b = \arg \min_{a,b} {\sum_i \left(\mathbb{V} [Y_i] - \frac{x_i}{a} - b^2\right)^2}.
    \label{eq:var}
\end{equation}

In our experiments, images have 8-bit depth, and thus
we only have 256 possible values for $x_i$. Hence, we can expect that sufficiently many pixels share the same intensity. However, this assumption limits this method, because it relies on images that have a few different pixel intensities, i.e., a sparse histogram, and that have a large dynamic range to get robust empirical variance values $\mathbb{V} [Y_i]$. Further, note that in Equation~\eqref{eq:var}, the same intensity level $x_i$ is appearing in the sum $\card{Y_i}$ times, and thus introduces a bias.

\subsection{Experimental results}
First, we provide results statistics of the Mean Squared Error (MSE) for the estimates ${\hat{a}}^{-1}, {\hat{b}}^2$ of the different methods compared to the ground truth values $a^{-1}, b^2$ in Tables~\ref{tab:stat_a} and~\ref{tab:stat_b}.
\begin{table}[ht]
    \caption{Statistics about the MSE error on ${\hat{a}}^{-1}$ for various methods.}
    \setlength\tabcolsep{3.5pt}
    \centering
    \begin{tabular}{ccccc}
        \toprule
        Method                 & Mean                    & Standard Dev.           & 75\%-Quantile                           & Maximum                 \\
        \midrule
        \textit{FOI}           & \num{3147.703950}       & \num{746292.063221}     & \num{0.000564}                          & \num{186467350.054470}  \\
        \textit{CNN}           & \num{0.017767}          & \num{0.086703}          & \num{0.000074}                          & \num{0.633754}          \\
        \textit{VAR}           & \num{0.000008}          & \num{0.000085}          & $\boldsymbol{\approx}$ \textbf{\num{0}} & \num{0.003535}          \\
        \textbf{\textit{OURS}} & \textbf{\num{0.000003}} & \textbf{\num{0.000014}} & \num{0.000001}                          & \textbf{\num{0.000277}} \\
        \bottomrule
    \end{tabular}
    \label{tab:stat_a}
\end{table}
\begin{table}[ht]
    \caption{Statistics about the MSE error on ${\hat{b}}^2$ for various methods.}
    \setlength\tabcolsep{3.5pt}
    \centering
    \begin{tabular}{ccccc}
        \toprule
        Method                 & Mean                                    & Standard Dev.           & 75\%-Quantile                           & Maximum                 \\
        \midrule
        \textit{FOI}           & \num{0.346023}                          & \num{6.674889}          & \num{0.000094}                          & \num{615.874927}        \\
        \textit{CNN}           & \num{0.000008}                          & \num{0.000023}          & \num{0.000005}                          & \num{0.000387}          \\
        \textit{VAR}           & \num{0.000001}                          & \num{0.000011}          & $\boldsymbol{\approx}$ \textbf{\num{0}} & \num{0.000445}          \\
        \textbf{\textit{OURS}} & $\boldsymbol{\approx}$ \textbf{\num{0}} & \textbf{\num{0.000001}} & $\boldsymbol{\approx}$ \textbf{\num{0}} & \textbf{\num{0.000033}} \\
        \bottomrule
    \end{tabular}
    \label{tab:stat_b}
\end{table}

In our experimental setting, \textit{FOI} performs worst compared to the other methods. This is expected, as the method only uses noisy observations $y$. That is the same for the \textit{CNN}, which, however, performs better. Thus, for most of the following graphs and plots, we only focus on our method \textbf{\textit{OURS}} compared with the baseline methods \textit{CNN} and \textit{VAR}.

\subsubsection{Mean squared error}
\begin{figure}[H]
    \centering
    \includegraphics[width=\columnwidth,trim={0 20 0 0},clip]{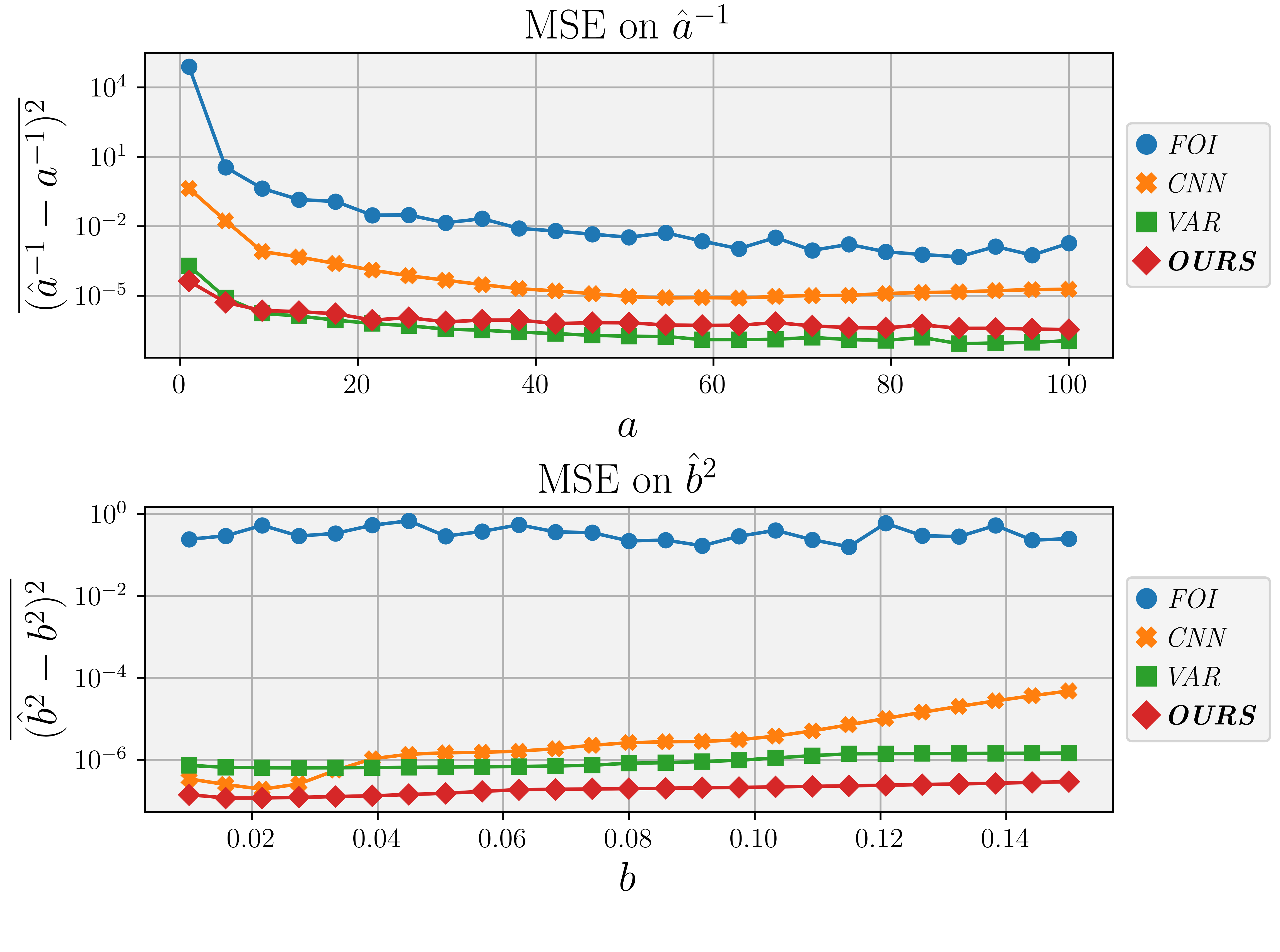}
    \caption{MSE for each method as a function of $a$ (top) and $b$ (bottom). Note that the error axis is in log scale.}
    \label{fig:mse_each_method}
\end{figure}

The MSE on ${\hat{a}}^{-1}$ is inversely correlated to the value of $a$, as shown in Fig.~\ref{fig:mse_each_method}. This fact can be explained by the dependence of our noise model on $a^{-1}$. For small $a$, the Poisson noise component is dominant. But when $a$ increases, the Poisson contribution to the noise model gradually vanishes. The MSE on ${\hat{b}}^2$ does not depend significantly on $b$, it is roughly constant for all methods except \textit{CNN}. Nonetheless, instances with less overall noise (large $a$ and small $b$) lead in general to smaller MSE values.

We note that our method consistently outperforms \textit{CNN}. Additionally, we also see that \textbf{\textit{OURS}} achieves a roughly $10$ times smaller MSE value on ${\hat{b}}^2$ than \textit{VAR}, while \textit{VAR} slightly improves MSE on ${\hat{a}}^{-1}$, particularly for larger $a$.

\subsubsection{Effect of outliers and image dependence}
\begin{figure}[H]
    \centering
    \includegraphics[width=\columnwidth,trim={0 20 0 0},clip]{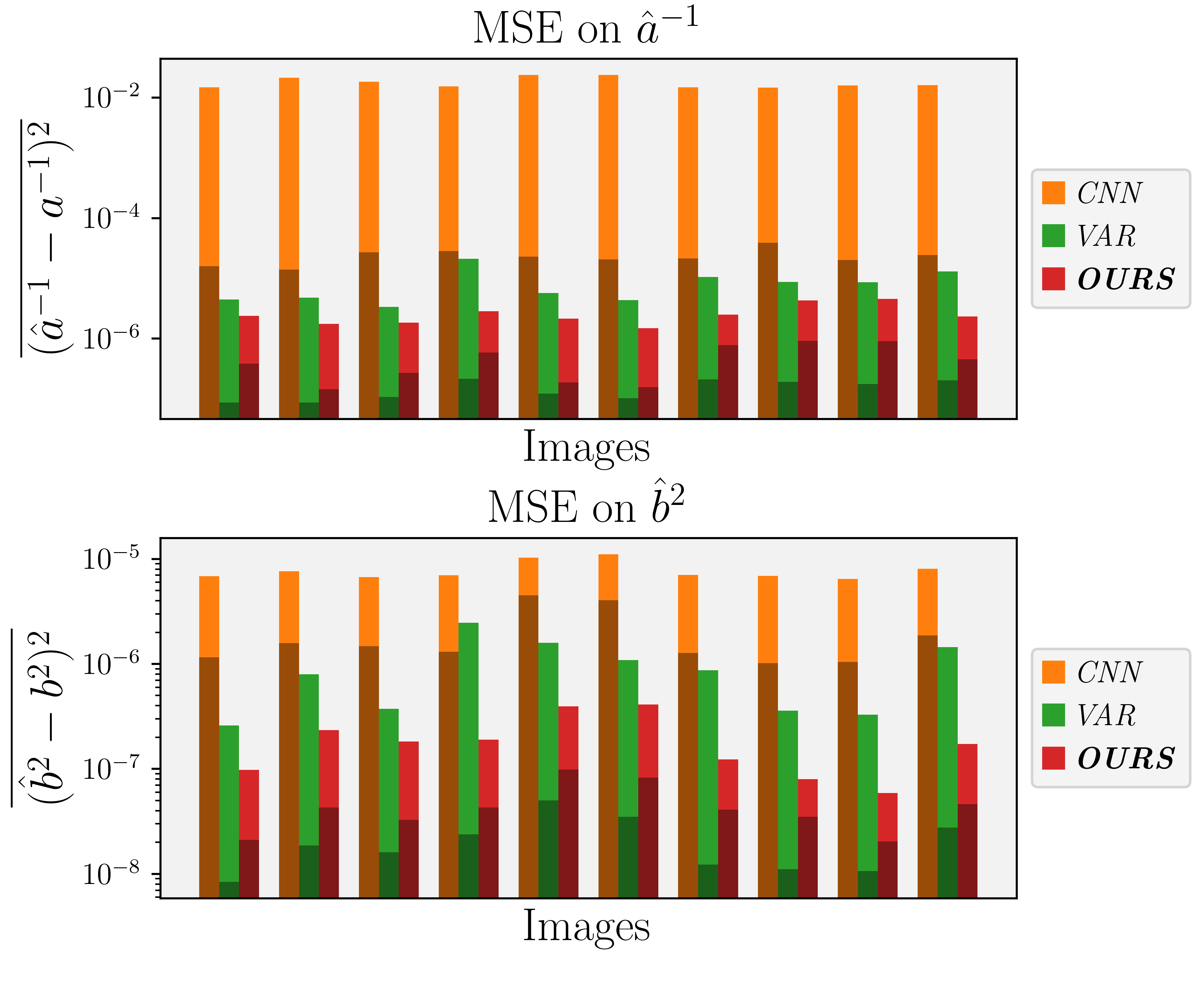}
    \caption{MSE dependence on $10$ images for ${\hat{a}}^{-1}$ (top) and ${\hat{b}}^2$ (bottom), including outliers (bright colors) and excluding outliers (dark colors). Note that the error axis is in log scale.}
    \label{fig:mse_dependence_image_outliers}
\end{figure}

We analyze the effect that outliers have on the overall performance of the different methods. We consider samples as outliers when $\text{MSE} > Q(0.75) + 1.5 * (Q(0.75) - Q(0.25))$ where $Q(0.25)$ and $Q(0.75)$ are the first and the third quartiles, respectively. We remove those outlier values, and Table~\ref{tab:out} shows the percentage of data remaining after filtering. In Fig.~\ref{fig:mse_dependence_image_outliers} we show the performance on 10 images~\cite{dataset:MartinFTM01}, with and without outliers. Excluding outliers significantly improves the performance.

\begin{table}[ht]
    \caption{Percentage of data not eliminated as outliers.}
    \setlength\tabcolsep{3.5pt}
    \centering
    \begin{tabular}{cccc}
        \toprule
        Method                 & $a$-based outliers                 & $b$-based outliers                 & Combined                           \\
        \midrule
        \textit{CNN}           & \percentage{0.821664}              & \percentage{0.841392}              & \percentage{0.701776}              \\
        \textit{VAR}           & \percentage{0.851600}              & \percentage{0.857357}              & \percentage{0.819280}              \\
        \textbf{\textit{OURS}} & \textbf{\boldpercentage{0.878112}} & \textbf{\boldpercentage{0.889520}} & \textbf{\boldpercentage{0.828320}} \\
        \bottomrule
    \end{tabular}
    \label{tab:out}
\end{table}

We can further observe that MSE varies depending on the intrinsic properties of the noise-free images $x$. We illustrate this dependence of the MSE in Fig.~\ref{fig:mse_dependence_image_outliers} by averaging over all different seeds and over the $a$ or $b$ values, respectively. One can observe that the ground truth image more significantly influences the estimation performance for the $b$ parameter. Additionally, we note that \textbf{\textit{OURS}} is the most robust when it comes to causing outlier error values, whereas \textit{CNN} is most prone to producing outliers.

\subsection{Analysis}

\subsubsection{Bias}
\begin{figure}[H]
    \centering
    \includegraphics[width=\columnwidth,trim={0 20 0 0},clip]{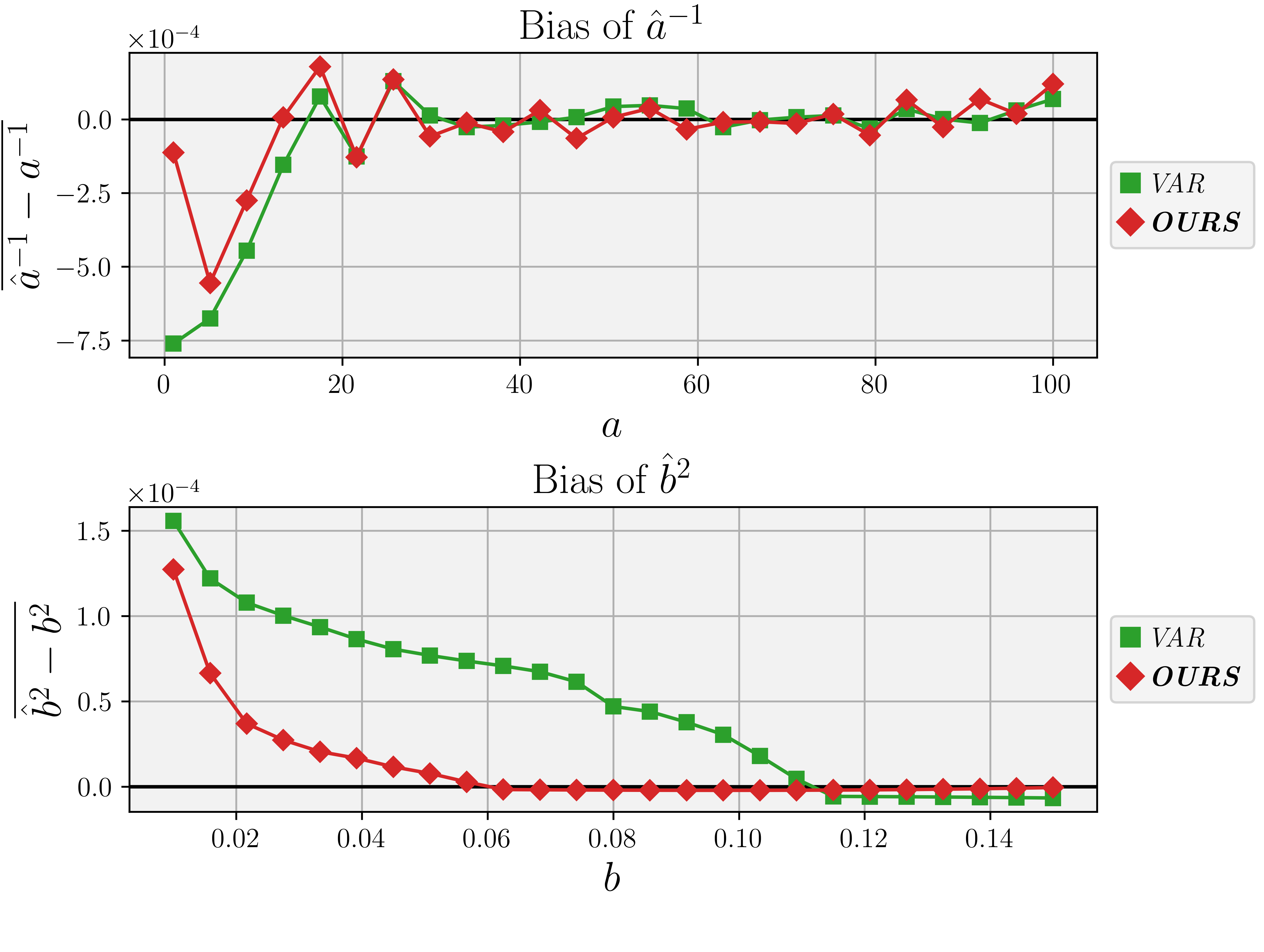}
    \caption{Absolute estimation error, illustrating the bias in ${\hat{a}}^{-1}$ (top) and ${\hat{b}}^2$ (bottom) of estimators \textit{VAR} and \textbf{\textit{OURS}}, for varying $a,b$ values. Note that the bias axis is in linear scale.}
    \label{fig:biasness_3_methods}
\end{figure}

Fig.~\ref{fig:biasness_3_methods} shows that the bias is most significant for both the smallest values of $a$ and of $b$, although the bias is small in general. For \textbf{\textit{OURS}} and ${\hat{a}}^{-1}$, this bias is explained by the variance of $\kappa_3[\mathcal{Y}]$ in Equation~\eqref{eq:cumu}, as this variance is dependent on $a^{-1}$ and is larger when $a$ is small.
Further, the bias on ${\hat{b}}^2$ comes from the fact that we only keep \textit{real} values of $\hat{b}$, discarding negative estimates of ${\hat{b}}^2$. By eliminating negative estimates, we introduce a positive bias. For \textit{VAR}, Equation~\eqref{eq:var} shows that under-estimating ${\hat{b}}^2$
leads to an over-estimation of ${\hat{a}}^{-1}$, which highlights the challenging un-mixing of the two noise parameters in this setup. Fig.~\ref{fig:biasness_3_methods}
also shows that \textit{VAR} is always biased in $b$ while for \textbf{\textit{OURS}} the bias is zero for $b > 0.06$.

\subsubsection{Log-likelihood} \label{sec:loglikelihood}

\begin{figure}[H]
    \centering
    \includegraphics[width=\columnwidth,trim={0 20 0 0},clip]{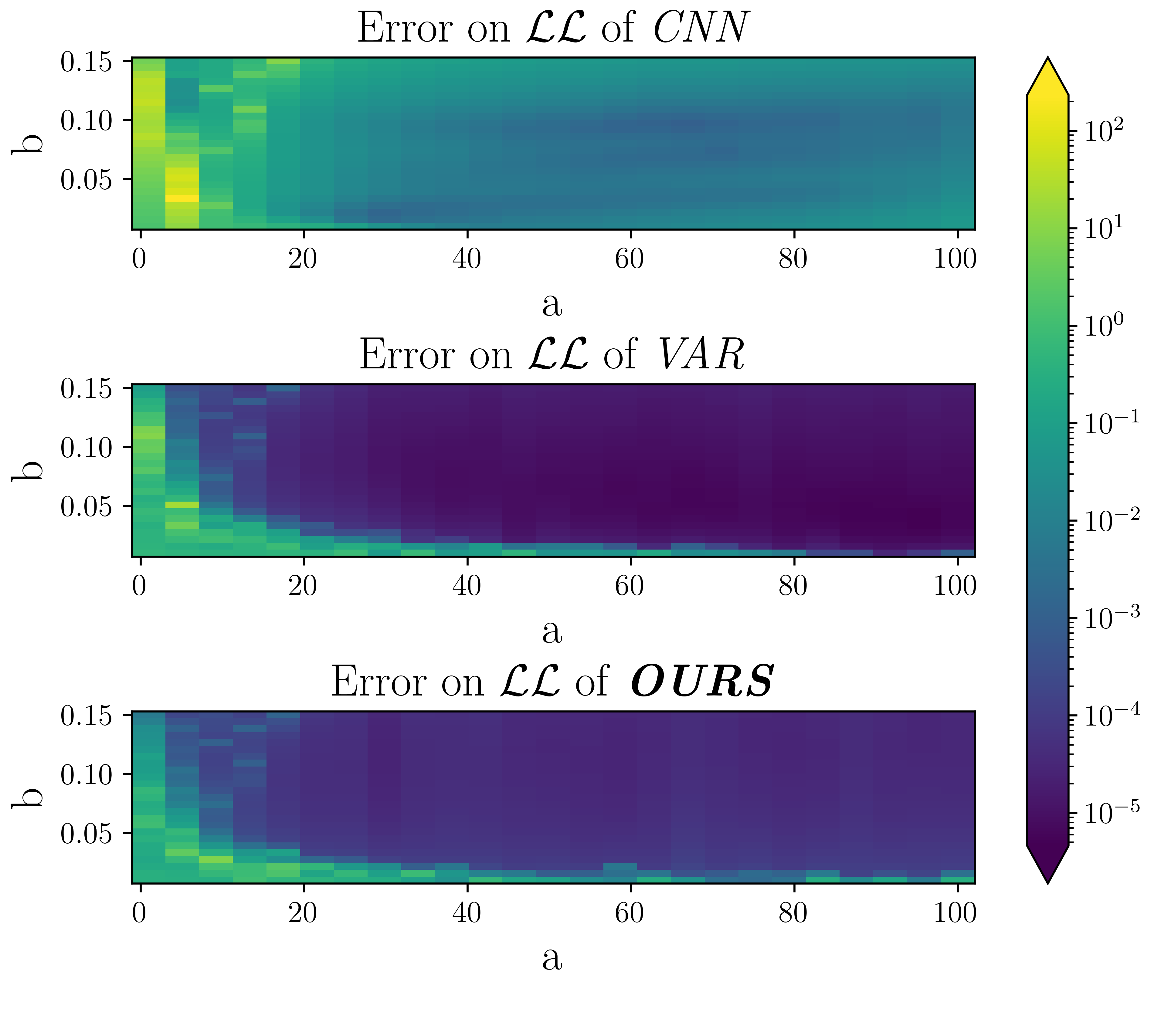}
    \caption{Relative absolute difference between the $\mathcal{LL}$ computed for the estimated parameters, and the actual $\mathcal{LL}$ of the ground truth parameters. Note that the error axis is in log scale.}
    \label{fig:mean_dll}
\end{figure}

As discussed in section~\ref{sec:loglikelihood}, $\mathcal{LL}$ cannot be efficiently optimized for global minima. However, empirically, $\mathcal{LL}$ can give additional insight. In Fig.~\ref{fig:mean_dll}, we show the relative absolute difference between $\mathcal{LL}$ for the actual values $a,b$ and their estimates $\hat{a},\hat{b}$ averaged over the validation images and seeds for different methods; $\overline{\left|\frac{\mathcal{LL}(y|\hat{a},\hat{b},x) - \mathcal{LL}(y|a,b,x)}{\mathcal{LL}(y|a,b,x)}\right|}$.

As shown in Fig.~\ref{fig:mean_dll}, \textit{CNN} leads to the biggest error. Moreover, \textit{VAR} results in estimates that are more “likely” on average, but, as shown earlier, performs worse than \textbf{\textit{OURS}}. This is due to the complexity of the statistical distribution of the noise model, leading to a mismatch between likelihood-maximization and expected-error-minimization estimators.

\subsubsection{Real-world scenario}
For real-world applications, the Poisson component often dominates the noise model ($a$ is small).
\textbf{\textit{OURS}} achieves smaller MSE than \textit{VAR} when $a$ is small. Therefore, while \textit{VAR} can provide more accurate $a$ estimates in certain cases, in real-world applications \textbf{\textit{OURS}} outperforms this baseline. Further, \textbf{\textit{OURS}} is more robust to outlier errors, is less biased, and consistently achieves smaller MSE on ${\hat{b}}^2$. \textit{VAR} also relies on images having a sparse histogram and a large dynamic range. For all these reasons, \textbf{\textit{OURS}} is better suited for real-world applications.

\section{Conclusion}\label{sec:ccl}
We propose an efficient cumulant-based Poisson-Gaussian noise estimator for paired noisy and noise-free images. Our method significantly outperforms prior baselines, notably a neural network solution and a variance-based method, both of which we design. Finally, the log-likelihood that we derive enables us to demonstrate the intrinsic difficulty of the Poisson-Gaussian noise estimation.

In future work, one could explore fine-tuning the weights of the \textit{VAR} baseline and taking into account the clipping behavior of digital sensors in real-world applications. Furthermore, we note that our method can be used as a starting point to speed up the optimization for maximizing the likelihood function that we derive, if likelihood —rather than inverse expected error— is to be maximized.

\bibliographystyle{IEEEtran}
\bibliography{paper}

\end{document}